\documentclass[letterpaper, 10 pt, journal, twoside]{ieeetran}
\usepackage{amsmath,amsfonts}
\usepackage{algpseudocode} 

\usepackage{algorithm}
\usepackage{array}
\usepackage[caption=false,font=normalsize,labelfont=sf,textfont=sf]{subfig}
\usepackage{textcomp}
\usepackage{stfloats}
\usepackage{url}
\usepackage{verbatim}
\usepackage{graphicx}
\usepackage{cite}
\usepackage{amssymb}
\usepackage{xspace}
\usepackage{booktabs}
\usepackage{tabularx}
\usepackage{multirow}
\usepackage{multicol}
\usepackage{adjustbox}
\usepackage{makecell}
\usepackage{colortbl}
\usepackage{xcolor}
\usepackage{textcomp}
\usepackage{algpseudocode}
\usepackage{hyperref}

\begin{document}

\title{\textit{CEI}: A Unified Interface for Cross-Embodiment Visuomotor Policy Learning in 3D Space}

\author{Tong Wu$^1$, Shoujie Li$^{12}$, Junhao Gong$^{1}$, Changqing Guo$^{1}$, Xingting Li$^{1}$, 
        Shilong Mu$^{3}$, and Wenbo Ding$^{1}$,~\IEEEmembership{Member,~IEEE}
\thanks{Manuscript received: September 26, 2025; Revised: December 6, 2025; Accepted: January 11, 2026.}
\thanks{This paper was recommended for publication by Editor Aleksandra Faust upon evaluation of the Associate Editor and Reviewers’ comments. 
This work was supported by National Key R\&D Program of China grant (2024YFB3816000), Guangdong Innovative and Entrepreneurial Research Team Program (2021ZT09L197), Shenzhen Science and Technology Program (JCYJ20220530143013030), Tsinghua Shenzhen International Graduate School-Shenzhen Pengrui Young Faculty Program of Shenzhen Pengrui Foundation (No. SZPR2023005) and Meituan. 
\textit{Tong Wu and Shoujie Li contributed equally to this
work.} (Corresponding author: Wenbo Ding, ding.wenbo@sz.tsinghua.edu.cn)}
\thanks{$^{1}$Tong Wu, Shoujie Li, Junhao Gong, Changqing Guo, Xingting Li, Wenbo Ding are with Shenzhen International Graduate School, Tsinghua University, Shenzhen 518055, China. (email: \{wu-t23, lsj20, gongjh24, gcq24, lixt25\}@mails.tsinghua.edu.cn, ding.wenbo@sz.tsinghua.edu.cn))}
\thanks{$^{2}$Shoujie Li is also with the School of Mechanical and Aerospace Engineering, Nanyang Technological University, Singapore 639956, Singapore. (email: shoujie.li@ntu.edu.sg)}
\thanks {$^{3}$Shilong Mu is with Xspark Ai, Shenzhen 518052, China. (email: mu.shilong@xspark-ai.com)}
\thanks{Digital Object Identifier (DOI): see top of this page.}
}




\maketitle

\definecolor{tablecolor}{HTML}{E0F4D7}
\newcommand{\CEI}{\textit{CEI}\xspace}
\newcommand{\ccbf}[1]{$\mathbf{#1}$}
\newcommand{\gray}[1]{\textcolor{gray}{$#1$}}

\begin{abstract}
Robotic foundation models trained on large-scale manipulation datasets have shown promise in learning generalist policies, but they often overfit to specific viewpoints, robot arms, and especially parallel-jaw grippers due to dataset biases. To address this limitation, we propose Cross-Embodiment Interface (\CEI), a framework for cross-embodiment learning that enables the transfer of demonstrations across different robot arm and end-effector morphologies. \CEI introduces the concept of \textit{functional similarity}, which is quantified using Directional Chamfer Distance. Then it aligns robot trajectories through gradient-based optimization, followed by synthesizing observations and actions for unseen robot arms and end-effectors. In experiments, \CEI transfers data and policies from a Franka Panda robot to \textbf{16} different embodiments across \textbf{3} tasks in simulation, and supports bidirectional transfer between a UR5+AG95 gripper robot and a UR5+Xhand robot across \textbf{6} real-world tasks, achieving an average transfer ratio of 82.4\%. Finally, we demonstrate that \CEI can also be extended with spatial generalization and multimodal motion generation capabilities using our proposed techniques. Project website: \url{https://cross-embodiment-interface.github.io/}.
\end{abstract}


\section{Introduction}
\label{sec:intro}

\IEEEPARstart{E}{merging} robotic foundation models are built upon scaling laws~\cite{lin2024data} and fueled by the growing availability of large-scale real-world manipulation datasets~\cite{Ebert-RSS-22, pmlr-v270-zhao25b}. However, these datasets often suffer from significant distributional imbalances, leading models to overfit to specific camera viewpoints and robot embodiments~\cite{pmlr-v270-chen25a}. For instance, OXE~\cite{o2024open} aggregates data from 60 datasets spanning multiple robotic platforms, yet remains heavily skewed toward Franka and xArm robots, with nearly all end-effectors limited to parallel grippers. Such biases limit the models’ ability to generalize~\cite{Gao-RSS-24}, especially when faced with embodiment variations.

To mitigate the embodiment biases, approaches such as Mirage~\cite{Chen-RSS-24} and RoVi-Aug~\cite{pmlr-v270-chen25a} employ techniques like cross-painting and generative models to synthesize visual observations, creating the illusion that the source robot is performing the task under test-time conditions. While these methods enable zero-shot deployment by bridging domain and embodiment gaps, their applicability is largely restricted to scenarios involving parallel-jaw grippers and Operational Space Control (OSC). These preconditions present significant challenges for transfer to more complex embodiments, such as multi-fingered dexterous hands. In fact, the limitation is further compounded by the scarcity of dexterous hand data~\cite{Wang-RSS-24}, which continues to hinder the development of generalizable policies across diverse robot embodiments.

\begin{figure}[t]
  \centering
  \includegraphics[width=0.95\linewidth]{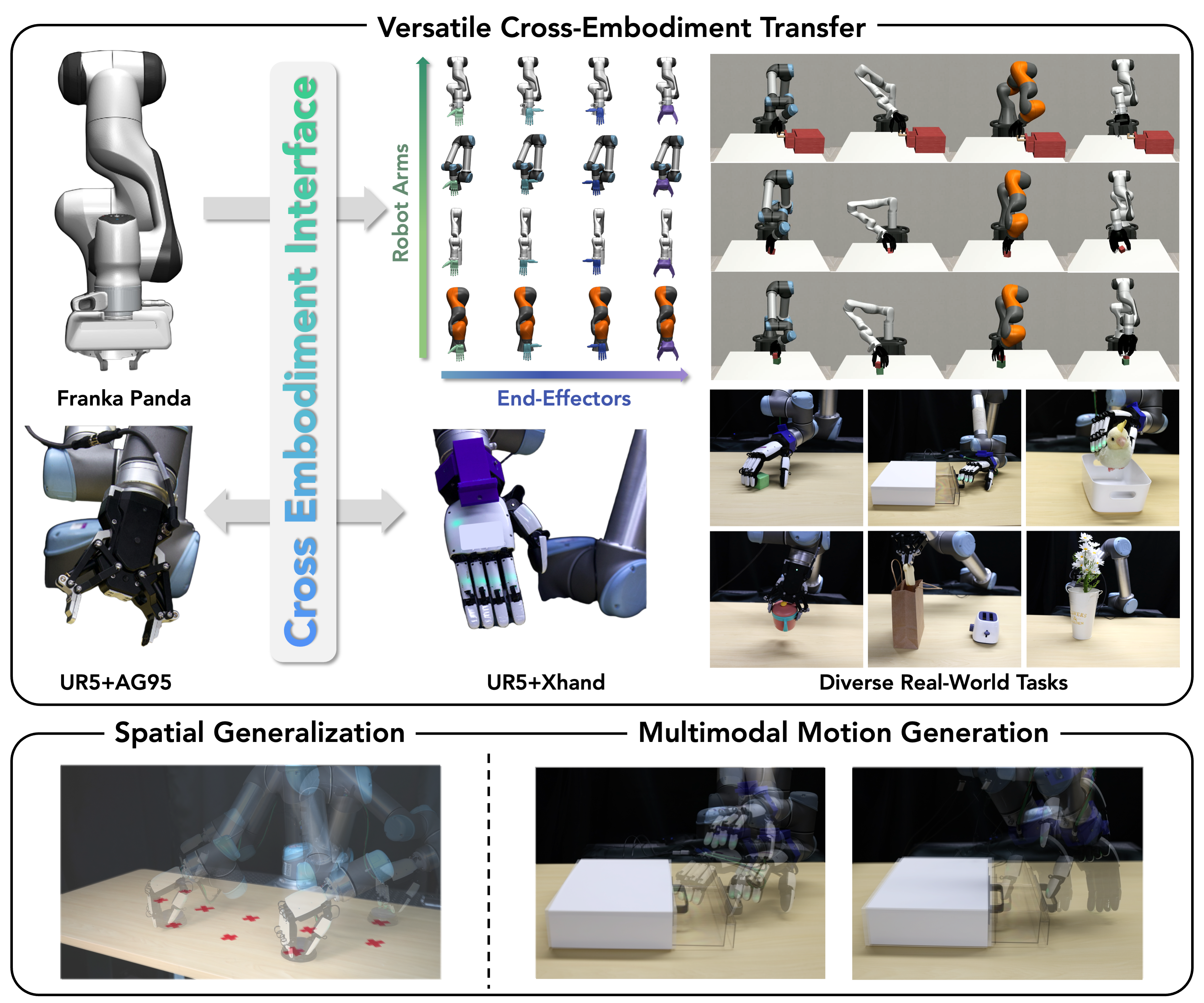}
  \caption{Cross-embodiment interface. \CEI enables cross-embodiment transfer between different robots by synthesizing demonstrations from a source embodiment to a target embodiment. We transfer data and policies from a Franka Panda robot to 16 target embodiments across 3 tasks in simulation, and demonstrate bidirectional transfer between a UR5+AG95 gripper and a UR5+Xhand setup across 6 real-world tasks. We also showcase \CEI's compatibility with spatial generalization and multimodal motion generation. }
  \label{fig:task}
  \vskip -0.1in
\end{figure}

Intriguingly, the underlying similarity in manipulation strategies between parallel-jaw grippers and dexterous hands suggests the feasibility of such cross-embodiment learning. For instance, when humans grasp a bottle, they often adopt a gripper-like pose that naturally promotes force closure. This motivates the question of whether such shared manipulation affordances can be systematically exploited to enable effective transfer across distinct end-effector morphologies.

To enable policy learning across heterogeneous robot embodiments, we propose Cross-Embodiment Interface (\CEI), a unified framework for cross-embodiment data synthesis. \CEI leverages a novel notion of \textit{functional similarity}, which captures shared interaction behaviors across different end-effectors, to align robot motions from a source embodiment to a target embodiment. This is accomplished by quantifying this similarity using the Directional Chamfer Distance~\cite{shi2024robocraft} between embodiments' functional representations, aligning trajectories via gradient-based optimization, and synthesizing corresponding observations and actions for the target robot. Through extensive experiments in both simulation and real world, we show that \CEI effectively transfers demonstrations and policies from a source robot equipped with a parallel-jaw gripper to a target robot with a five-fingered dexterous hand and vice versa, with an overall transfer ratio of 82.4\%. Furthermore, we demonstrate that \CEI is compatible with spatial generalization and support multimodal data generation using our extended techniques.

Our contributions are summarized as follows:
\begin{itemize}
    \item We propose a novel concept of \textit{functional similarity} based on Directional Chamfer Distance, coupled with gradient-based trajectory alignment, to transfer task-relevant manipulation behaviors across embodiments.

    \item We propose a general pipeline that leverages embodiment information to augment both observations and actions, enabling the synthesis of point cloud–based demonstrations across heterogeneous robots.

    \item Experiments on 16 embodiments over 3 simulation tasks and bidirectional transfers over 6 real-world tasks demonstrate \CEI's effectiveness on cross-embodiment learning. Furthermore, we showcase that \CEI can be extended with spatial generalization and multimodal motion generation for diverse data synthesis.
\end{itemize}

\section{Related Work}
\label{sec:rw}

\subsection{Data Generation for Robotic Manipulation} 
While recent imitation learning methods have exhibited impressive performance~\cite{chi2023diffusion, Ze2024DP3}, the high cost of collecting expert demonstrations poses significant challenges to scalability and real-world deployment. To mitigate this, online data generation approaches adapt existing demonstrations to novel object configurations and produce plausible interaction trajectories via rollouts in high-fidelity physics simulators~\cite{pmlr-v229-mandlekar23a}. Although these methods preserve physical realism, they tend to be computationally intensive, and challenging to apply directly in real-world settings.
In contrast, offline data generation synthesizes new demonstrations from existing datasets through trajectory transformations~\cite{xue2025demogen} or generative model-based visual augmentations~\cite{chen2023genaug}. 
\CEI focuses on cross-embodiment transfer through offline data generation, but addresses the challenge of extreme differences between end-effectors, enabling the learning of dexterous hand policies from parallel gripper data.
 
\subsection{Cross-embodiment Learning} 
Prior work has explored bridging the embodiment gap through various strategies, including space alignment~\cite{ZhangXEPW21, yuan2024cross, Bauer2025LatentAD}, cross-painting~\cite{Chen-RSS-24, pmlr-v270-chen25a}, dynamics modeling~\cite{HuHRJ22}, and reward model learning~\cite{pmlr-v164-zakka22a}. Other approaches incorporate embodiment information directly into the policy and train across a set of embodiments with varying kinematics and dynamics, demonstrating generalization to unseen morphologies within the training distribution~\cite{furuta2023asystem, shao2020unigrasp,xu2021adagrasp}. 
Recently, many efforts have focused on collecting cross-embodiment datasets~\cite{Ebert-RSS-22, o2024open, pmlr-v229-walke23a}, which have been shown to generalize effectively across embodiments~\cite{Yang-RSS-24, octo_2023}. In contrast to these efforts, our work addresses extreme cross-embodiment data generation, specifically demonstration transfer between a  parallel gripper and a dexterous hand, without requiring embodiment-specific training data. 

\section{Problem Formulation}
A visuomotor policy $\pi : \mathcal{O} \rightarrow \mathcal{A}$ maps visual observations $o \in \mathcal{O}$ to actions $a \in \mathcal{A}$. In imitation learning, such policies are typically trained from a dataset of demonstrations $\mathcal{D}$ collected on a particular embodiment $E$.
We represent each demonstration $D_{E, s_0} \subseteq \mathcal{D}$ as a trajectory of observation-action pairs, conditioned on an initial task state $s_0$ and robot embodiment $E$: $
D_{E, s_0} = \left((o_0, a_0), (o_1, a_1), \dots, (o_{L-1}, a_{L-1}) \mid s_0, E \right).$ 
Each observation $o_t = (o_t^{\text{pcd}}, o_t^{\text{arm}}, o_t^{\text{ee}})$ consists of a point cloud and the proprioceptive states of the robot arm and end-effector, while each action $a_t = (a_t^{\text{arm}}, a_t^{\text{ee}})$ specifies joint position targets for robot arm and end-effector. 
Our objective is to generate a corresponding demonstration $\hat{D}_{E', s_0}$ for a new target embodiment $E'$ potentially differing in morphology and kinematics, starting from the same initial task state $s_0$: $
\hat{D}_{E', s_0} = \left((\hat{o}_0, \hat{a}_0), (\hat{o}_1, \hat{a}_1), \dots, (\hat{o}_{L-1}, \hat{a}_{L-1}) \mid s_0, E' \right).$ 
Unlike prior work that focuses solely on the initial state or short trajectory segments, we aim to produce full demonstration trajectories that are executable on the target embodiment.
Each embodiment $E$ includes structural specifications such as joint limits, robot meshes and kinematic chains. We assume access to standardized robot description files (e.g., URDF and XML), which provide this information and allow embodiment-specific adaptation of actions and observations.

\section{Method}
\label{sec:method}
\begin{figure*}[]
  \centering
  \vspace{2pt}
  \includegraphics[width=0.9\linewidth]{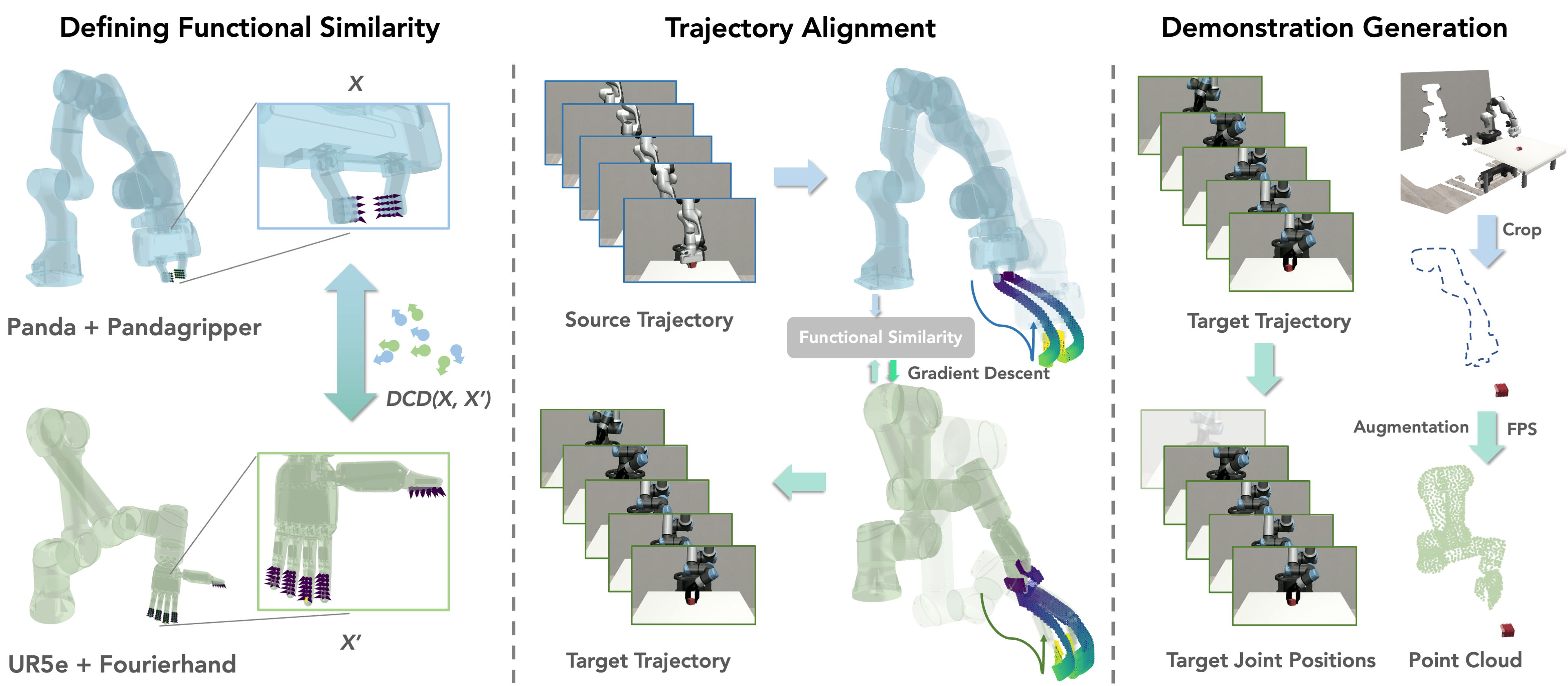}
  \caption{Overview of the pipeline. Given a source dataset, a source embodiment, and a target embodiment, we first define functional representations as sets of points with associated directions on both embodiments. We then compute functional similarity using the negative Directional Chamfer Distance between these representations. Trajectory alignment is performed by sequentially optimizing the functional similarity for each trajectory slice. Finally, we synthesize target actions with next joint positions and generate target observations by augmenting source point clouds with points sampled from the target embodiment. The viridis colormap is used to illustrate the temporal progression of the trajectory of functional representations. }
  \label{fig:pipeline}
  \vskip -0.1in
\end{figure*}

To generate demonstrations for a new embodiment, \CEI first defines the functional representations of both the source and target embodiments and employs the Directional Chamfer Distance to quantify the \textit{functional similarity} between the two embodiments (Section~\ref{sec:func_sim}). Leveraging the source demonstrations, functional representations and the metric, \CEI aligns the robot trajectories from the source embodiment to the target embodiment (Section~\ref{sec:traj_align}). Finally, \CEI synthesizes corresponding observations and actions based on the aligned trajectories, obtaining the demonstrations for the target embodiment (Section~\ref{sec:oa_generation}). An overview of the full pipeline is illustrated in Fig.~\ref{fig:pipeline}.

\subsection{Functional Similarity between Embodiments}
\label{sec:func_sim}

While two embodiments may differ significantly in morphology and kinematics, they can nonetheless exhibit similar object interaction behaviors. For example, in a pick-up task, a dexterous hand may grasp an object from opposing sides using the thumb and other four fingers—a strategy that is functionally analogous to the symmetric grasp of a parallel gripper. In this section, we propose a method to formally quantify this \textit{functional similarity} between embodiments.

\subsubsection{Functional Representation}
We represent an embodiment’s interaction feature through a set of point-direction pairs $X=\{(p_i, n_i)\}_{i=0,1,...,N}$,  referred to as the functional representation. As shown in Fig.~\ref{fig:pipeline}, the points \(\{p_i\} \) lie on the embodiment’s surface and reflect potential contact areas with objects. The associated directions \( \{n_i\} \) are automatically computed as the surface normal in the vicinity of \( \{p_i\} \). All point-direction pairs are transformed to the world frame via forward kinematics, ensuring alignment happens in a common frame across embodiments. While this concept is inspired by \textit{affordance}~\cite{bahl2023} which captures potential interaction sites on the object, we instead focus on the embodiment, emphasizing how the end-effector interacts with the environment.

\subsubsection{Directional Chamfer Distance}  
To quantify the similarity between two functional representations, we adopt the \textit{Directional Chamfer Distance} (DCD), which jointly considers spatial proximity and directional alignment between point sets. Specifically, given functional representations \( X = \{(p_i, n_i)\}_{i=1}^{N} \) and \( X' = \{(p'_j, n'_j)\}_{j=1}^{N'} \) for the source and target embodiments respectively, DCD is defined as:
\begin{equation}
\begin{aligned}
    \mathrm{DCD}(X, X') &=
    \frac{1}{N} \sum_{i=1}^{N} \min_{j} \left( \|p_i\! -\! p'_j\|_2 \!-\! \lambda\! \cdot\! \langle n_i, n'_j \rangle \right)\\
    &+\frac{1}{N'} \sum_{j=1}^{N'} \min_{i} \left( \|p'_j\! -\! p_i\|_2 \!-\! \lambda \!\cdot \!\langle n'_j, n_i \rangle \right),
\end{aligned}
\end{equation}
where \( \lambda \) is a weighting factor balancing spatial and directional terms. This formulation ensures that each point-direction pair in one set is matched to its most similar counterpart in the other and calculates the distance of two sets. We define the negative DCD as the measure of \textit{functional similarity}, which increases as the functional representations of two embodiments become more consistent.

\subsection{Trajectory Alignment}
\label{sec:traj_align}

To bridge the embodiment gap, we leverage the defined \textit{functional similarity} to align the joint (proprioceptive) trajectories between two robots. Given a source embodiment and an associated demonstration, we first compute the trajectory of its functional representation \( \{X_t\}_{t=0}^{L-1} \) using a differentiable forward kinematics module \text{FK}, where \( X_t = \text{FK}_E(o_t^{\text{arm}}, o_t^{\text{ee}}) \).
For each frame \( t \), we initialize the target embodiment’s joint configuration \( q'_t  \) as a set of learnable variables and calculate the corresponding functional representation \( X'_t = \text{FK}_{E'}(q'_t) \).
We then optimize functional similarity between \( X_t \) and \( X'_t \) using gradient descent:
\begin{equation}
\min_{q'_t} \mathcal{L}_{\text{align}}(q'_t) = w_1 \mathrm{DCD}(X_t, X'_t) + w_2 \mathcal{L}(q'_t),
\end{equation}
where \( \mathcal{L}(q'_t) \) penalizes the out-of-range joint configuration, and $w_1$, $w_2$ are weighting factors.
Rather than optimizing each frame independently, we treat the process as a sequential optimization problem. Specifically, the joint configuration at frame \( t+1 \) is initialized using the optimized result from frame \( t \): \( q^{\prime(0)}_{t+1} \leftarrow \hat{q}_{t}, \) where \( q^{\prime(0)}_{t+1} \) represents the initial joint configuration at \( t+1 \) and \( \hat{q}_{t} \) denotes the optimized joint configuration at \( t \). This strategy not only accelerates convergence due to the small variation between adjacent frames, but also ensures temporal consistency in the generated trajectory.

\subsection{Observation and Action Generation}
\label{sec:oa_generation}

\subsubsection{Action generation}  
We define the action at each timestep as the target embodiment's joint configuration at the next frame: \( a_t = \hat{q}_{t+1}. \)
Since the embodiment may not achieve the target position in time, we adopt a closed-loop control strategy during execution, where the embodiment continuously applies \( a_t \) until its current joint configuration reaches the target position. 

\subsubsection{Observation generation}  
The proprioceptive observation of the target embodiment is directly derived from the aligned trajectory: \( (\hat{o}_t^{\text{arm}}, \hat{o}_t^{\text{ee}}) = \hat{q}_t. \)
\textcolor{black}{To synthesize the point cloud, we first remove points that lie outside the defined workspace, and mask any point that falls within a distance threshold $\tau$ (e.g., 5 mm) of the source robot. This proximity is calculated using the minimum Euclidean distance to a uniformly sampled point cloud derived from the robot's mesh. Next, we synthesize the point cloud for the target embodiment by first accessing target robot description file and then sampling points across the robot's mesh. Both the cropping and augmentation steps are conditioned on the current proprioceptive state \( \hat{q}_t \). Finally we apply Farthest Point Sampling (FPS) to downsample the point cloud to a fixed number (1024).} During inference, the same process is applied to ensure consistency between the synthesized observations and those encountered at test time.

\section{Experimental Setup}
\label{sec:exp}

\begin{figure}[t]
  \centering
  \vspace{4pt}
  \includegraphics[width=\linewidth]{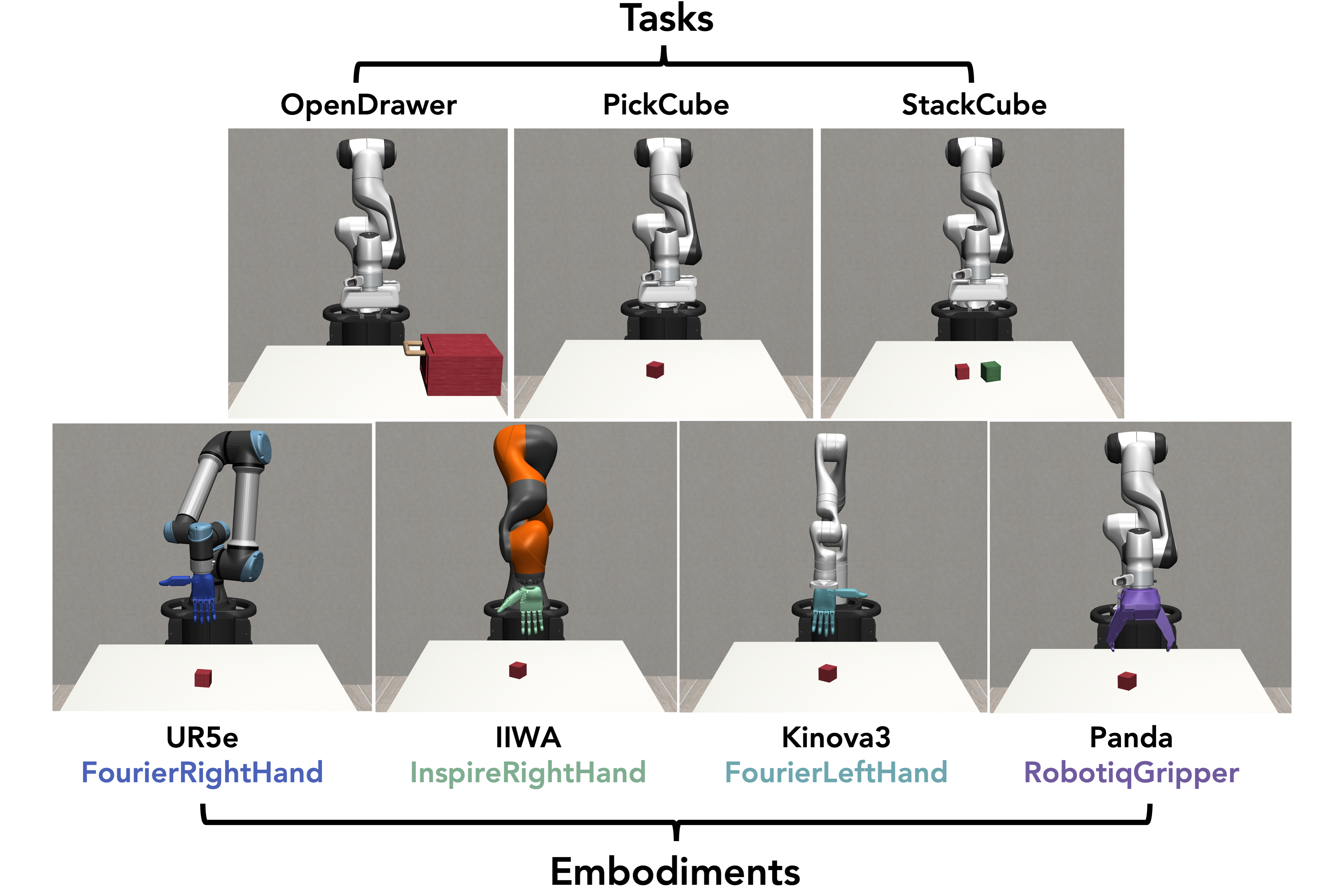}
  \caption{Tasks and embodiments for simulation evaluation. We investigate 3 tasks and 16 embodiments, which are combinations of 4 robot arms (UR5e, IIWA, Kinova3 and Franka Panda) and 4 end-effectors (FourierRighthand, InspireRightHand, FourierLefthand and RobotiqThreeFinger gripper). }
  \label{fig:task}
  \vskip -0.1in
\end{figure}

\subsection{Implementation Details}
\label{sec:implementation}

To construct the robot description, we utilize the \textit{XML} files of the robot arms and end-effectors. The points of functional representation are sampled from the finger pad meshes, and their associated directional vectors are generated using \textit{TorchSDF}. For efficient trajectory alignment, we employ \textit{pytorch\_kinematics} as a batch forward kinematics engine, enabling highly parallelized computations.
Optimization is performed with a maximum of 300 steps at each timestep. Early stopping is triggered if the alignment loss \( \mathcal{L}_{\text{align}} \) does not improve for 10 consecutive steps. We empirically determine that setting the weighting factor \(\lambda = 0.5\), and weights \(w_1=1\), \(w_2=1\) yields robust performance across all tasks.

\subsection{Policy Training}
\label{sec:policy}

We utilize 3D Diffusion Policy~\cite{Ze2024DP3} to evaluate the synthesized data. The input consists of the current joint positions of both the robot arm and end-effector, along with a preprocessed point cloud of size [1024, 3]. The policy outputs target joint positions. We set the observation horizon to $T_o = 2$, action prediction horizon to $T_p = 16$, and action execution horizon to $T_a = 8$, following the settings in~\cite{Ze2024DP3}. The model is trained for 3000 epochs using the AdamW optimizer with a learning rate of $1 \times 10^{-4}$ and a 500-step warmup for training stabilization. All experiments are conducted on a single RTX 4090 GPU.

\subsection{Evaluation Setup}

\subsubsection{Simulation}
We evaluate cross-embodiment transfer and visuomotor policy learning in simulation using \textbf{3} manipulation tasks and \textbf{16} robot embodiments (combinations of \textbf{4} arms and \textbf{4} end-effectors) from robosuite~\cite{Zhu2020robosuiteAM} (as shown in Fig.~\ref{fig:task}). Demonstrations collected on a Franka Panda via teleoperation are transferred to target embodiments using \CEI. We evaluate transferred trajectories via simulation replay and train DP3 policies on the synthesized data, measuring performance over 20 trials with three seeds.


\begin{figure}[t]
  \centering
  \vspace{4pt}
  \includegraphics[width=0.95\linewidth]{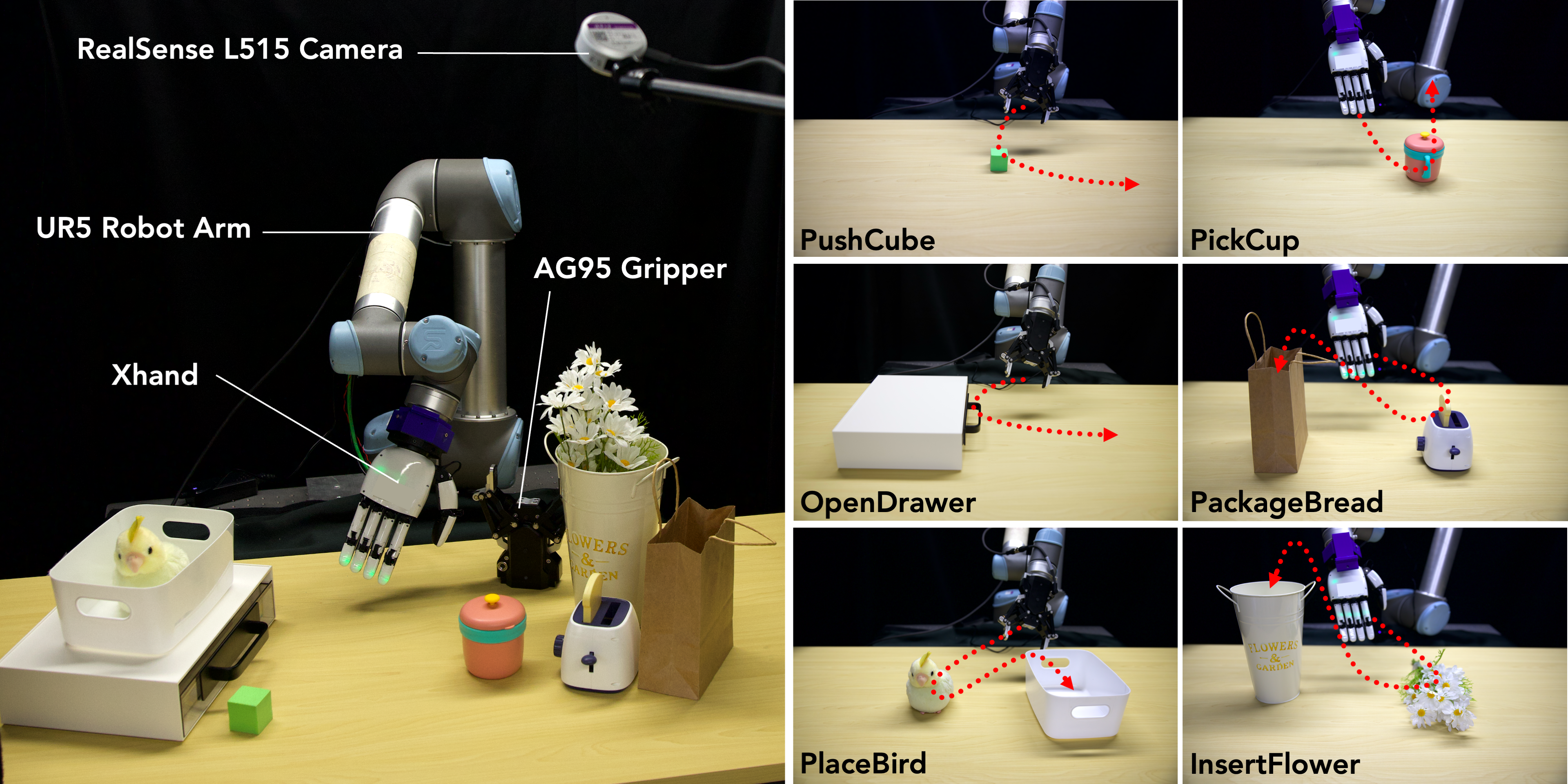}
  \caption{Left: Setup and associated objects in real-world experiments. Right: Real-world tasks. We evaluate transfer from the AG95 gripper to the Xhand on PushCube, OpenDrawer, and PlaceBird, and from the Xhand to the AG95 gripper on PickCup, PackageBread, and InsertFlower.}
  \label{fig:real_tasks}
  \vskip -0.1in
\end{figure}

\begin{table*}[!t]
\centering
\vspace{6pt}
\caption{Full Evaluation across Tasks and Embodiment Combinations in Simulation. }
\label{tab:full_eval}
\begin{tabular}{c|cccc|cccc|cccc}
\toprule
\textbf{Tasks} & \multicolumn{4}{c|}{OpenDrawer} & \multicolumn{4}{c|}{PickCube} & \multicolumn{4}{c}{StackCube} \\
\midrule
\textbf{Grippers / Robot Arms} & UR5e & IIWA & Kinova3 & Panda & UR5e & IIWA & Kinova3 & Panda & UR5e & IIWA & Kinova3 & Panda \\
\midrule
FourierRightHand     & \(100\) & \(100\) & \(100\) & \(100\) & \(81\) & \(72\) & \(69\) & \(79\) & \(53\) & \(38\) & \(37\) & \(34\) \\
InspireRightHand     & \(91\)  & \(87\)  & \(87\)  & \(84\)  & \(50\) & \(45\) & \(33\) & \(43\) & \(24\) & \(16\) & \(12\) & \(15\) \\
FourierLeftHand      & \(100\) & \(96\)  & \(98\)  & \(100\) & \(88\) & \(83\) & \(81\) & \(79\) & \(57\) & \(47\) & \(50\) & \(47\) \\
RobotiqThreeFinger   & \(78\)  & \(91\)  & \(93\)  & \(93\)  & \(19\) & \(31\) & \(41\) & \(52\) & \(0\)  & \(0\)  & \(5\)  & \(9\)  \\
\bottomrule
\end{tabular}
\vskip -0.1in
\end{table*}

\subsubsection{Real world}
In real world deployment, demonstrations are collected via keyboard teleoperation on a UR5 arm, with RGB-D data captured by a RealSense L515 to generate scene point clouds, as shown in Fig.~\ref{fig:real_tasks} (left). DP3 policies are trained on \CEI-generated data and evaluated over 10 trials per task. We evaluate transfer between the AG95 gripper and Xhand on \textbf{6} real-world tasks (Fig.~\ref{fig:real_tasks} right): 
\begin{itemize}
    \item \textbf{PushCube.} A cube is placed on the table. The robot must approach and align its end-effector to securely enclose the cube, then push it 20 cm to the right.
    
    \item \textbf{OpenDrawer.} A drawer is  positioned on the left side of the table. The robot moves toward the handle, inserts its fingertip, and pulls it to the right by about 10 cm.
    
    \item \textbf{PlaceBird.} A toy bird is  positioned on the left side of the table, while a box is fixed on the right. The robot approaches, grasps the bird stably, moves it over the box, lowers it, and places the bird inside.
    
    \item \textbf{PickCup.} A cup is placed near the center of the table. The robot approaches, grasps the cup, and lifts it about 5 cm from the surface.

    \item \textbf{PackageBread.} A toaster with baked bread is placed on the right side of the table, with a bag fixed on the left. The robot approaches the toaster, picks up the bread, moves it above the bag, and deposits the bread inside.

    \item \textbf{InsertFlower.} A bouquet of flowers is placed on the right side of the table, and a vase is fixed on the left. The robot approaches, grasps the bouquet, rotates and moves it above the vase, then inserts and releases it.

\end{itemize}
For the first three tasks, we collect 25 AG95 demonstrations and transfer to Xhand; for the latter three, we collect data with Xhand and transfer to AG95.

\section{Results}

\subsection{Simulation Results}

\subsubsection{Gripper-to-hands transfer} 

To evaluate whether CEI effectively bridges the extreme embodiment gap from parallel-jaw grippers to dexterous multi-fingered hands, we validate the synthesized data by replaying the robot trajectory online, initialized with the same state when collecting the data.
Table~\ref{tab:full_eval} presents the comprehensive evaluation of 16 different embodiments. The results indicate that despite variations in kinematics and morphology, \CEI is capable of bridging the cross-embodiment gap by leveraging \textit{functional similarity}. 
\textcolor{black}{We further observe that transfer difficulty correlates with contact richness. While OpenDrawer involves simple interactions, PickCube requires stable grasp acquisition, and StackCube entails sequential contact maintenance. These contact-rich scenarios amplify the susceptibility of our geometry-based synthesis to physical disturbances (e.g., slippage), explaining the observed performance degradation.}
Analyzing performance across end-effectors, we find that the FourierRightHand and FourierLeftHand consistently achieve the highest success rates across all three tasks. In contrast, the InspireRightHand experiences an average performance drop of approximately 40\%. 
\textcolor{black}{Although the RobotiqThreeFinger exhibits minimal loss in OpenDrawer, it struggles with StackCube as kinematic constraints force a transition from stable `finger-pad' grasping to a low-contact `fingertip' strategy, thereby reducing manipulation reliability.}

\subsubsection{Ablation study on cross-embodiment techniques}

Table~\ref{tab:online_exp} presents an ablation study to investigate the key components that enable \CEI to achieve cross-embodiment transfer. We compare \CEI against two baselines: (1) Binary Manual Specification (BMS) generates end-effector motions by linearly interpolating between manually specified open and close poses, while constraining the end-effector to match the gripper aperture of the source embodiment, and (2) \CEI without Direction removes directional information and uses only positional features for functional representation. We use a subset of embodiments: Emb. 1 = UR5e + FourierRightHand, Emb. 2 = IIWA + InspireRightHand, Emb. 3 = Kinova3 + FourierLeftHand, and Emb. 4 = Panda + RobotiqThreeFinger.
The results show that \CEI without Direction achieves an average success rate of only 32\%, only half of \CEI. Although it is still capable of completing the OpenDrawer task, it struggles with object grasping tasks, highlighting the critical role of directional information for grasp motions. \textcolor{black}{We also observed that BMS failed in PickCube and StackCube because linear interpolation between manually defined poses often yields failed grasps. Additionally, kinematic discrepancies (e.g., end-effector frame offsets) caused frequent failures in OpenDrawer, rendering BMS inferior to \CEI even when the degree of opening were explicitly constrained to be the same.}

\begin{table}[ht]
\centering
\scriptsize
\setlength{\tabcolsep}{4pt}

\caption{Ablation Study on Cross-embodiment Techniques.}
\label{tab:online_exp}
\begin{tabular}{llccccc}
\toprule
\textbf{Task} & \textbf{Method} & Emb. 1 & Emb. 2 & Emb. 3 & Emb. 4 & \textbf{Avg.} \\
\midrule

\multirow{3}{*}{\textit{OpenDrawer}} 
 & BMS                & \(80\) & \(92\) & \(87\) & \(64\) & \(81\) \\
 & \CEI w/o Dir.      & \(84\) & \(\boldsymbol{100}\) & \(91\) & \(\boldsymbol{100}\) & \(94\) \\ 
 & \textbf{\CEI (Ours)} & \(\boldsymbol{100}\) & \(87\) & \(\boldsymbol{98}\) & \(93\) & \(\boldsymbol{95}\) \\
\midrule

\multirow{3}{*}{\textit{PickCube}} 
 & BMS                & \(0\) & \(0\) & \(0\) & \(0\) & \(0\) \\
 & \CEI w/o Dir.      & \(7\) & \(2\) & \(0\) & \(0\) & \(2\) \\
 & \textbf{\CEI (Ours)} & \(\boldsymbol{81}\) & \(\boldsymbol{45}\) & \(\boldsymbol{81}\) & \(\boldsymbol{52}\) & \(\boldsymbol{65}\) \\
\midrule

\multirow{3}{*}{\textit{StackCube}} 
 & BMS                & \(0\) & \(0\) & \(0\) & \(0\) & \(0\) \\
 & \CEI w/o Dir.      & \(0\) & \(0\) & \(0\) & \(0\) & \(0\) \\
 & \textbf{\CEI (Ours)} & \(\boldsymbol{53}\) & \(\boldsymbol{16}\) & \(\boldsymbol{50}\) & \(\boldsymbol{9}\) & \(\boldsymbol{32}\) \\

\bottomrule
\end{tabular}
\end{table}

\subsubsection{Sensitivity of functional representation}

To evaluate the impact of different functional representation selections on transfer performance, we compare three variants: (1) Standard, which spans the entire finger pad to provide full coverage and is used across main experiments; (2) Reduced, which is limited to a subset of points near the center of the finger pad; and (3) Randomly Dropped (Rand. Drop.), where a subset of points is randomly removed from the full representation. We transfer 20 demonstrations with above three selections in the PickCube task using FourierRightHand and evaluate the success rate. As shown in Table~\ref{tab:fr_sensitivity}, although the three selections differ, their success rates remain comparable, suggesting that \CEI is robust to such variations and exhibits low sensitivity to the choice of functional representation.

\begin{table}[ht]
\caption{Sensitivity Analysis of Functional Representations.}
\centering
\scriptsize
\begin{tabular}{cccc}
\toprule
\makecell{Functional\\Representation} & 
\raisebox{-0.\height}{\includegraphics[width=0.030\textwidth]{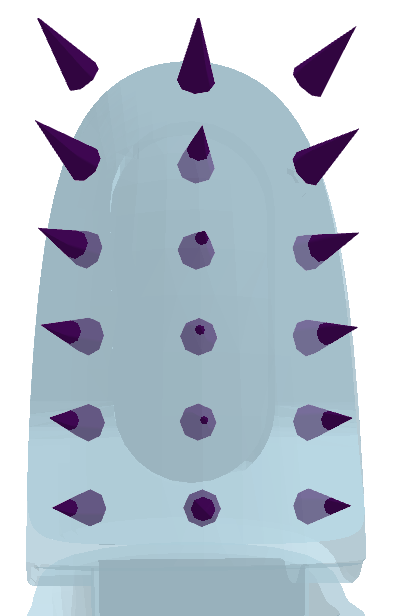}} & 
\raisebox{-0.\height}{\includegraphics[width=0.030\textwidth]{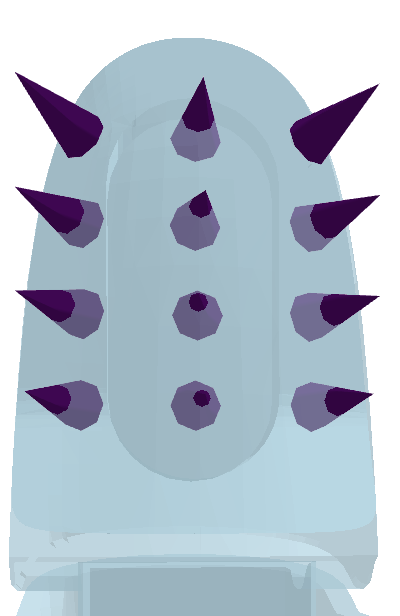}} & 
\raisebox{-0.\height}{\includegraphics[width=0.030\textwidth]{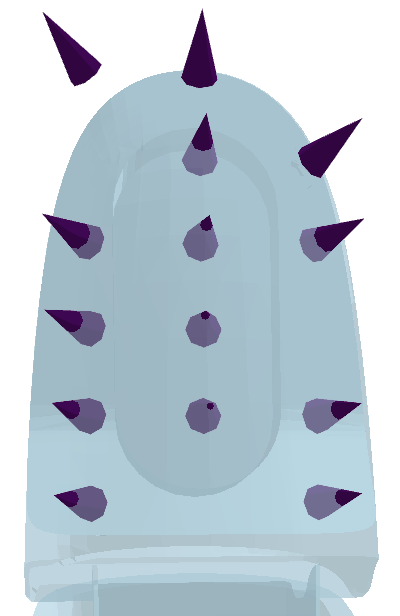}} \\
& Standard & Reduced & Rand. Drop. \\
\midrule
Success Rate & \(17\)/\(20\) & \(18\)/\(20\) & \(16\)/\(20\) \\
\bottomrule
\end{tabular}
\label{tab:fr_sensitivity}
\end{table}

\begin{figure*}[t]
  \centering
  \vspace{4pt}
  \includegraphics[width=0.95\linewidth]{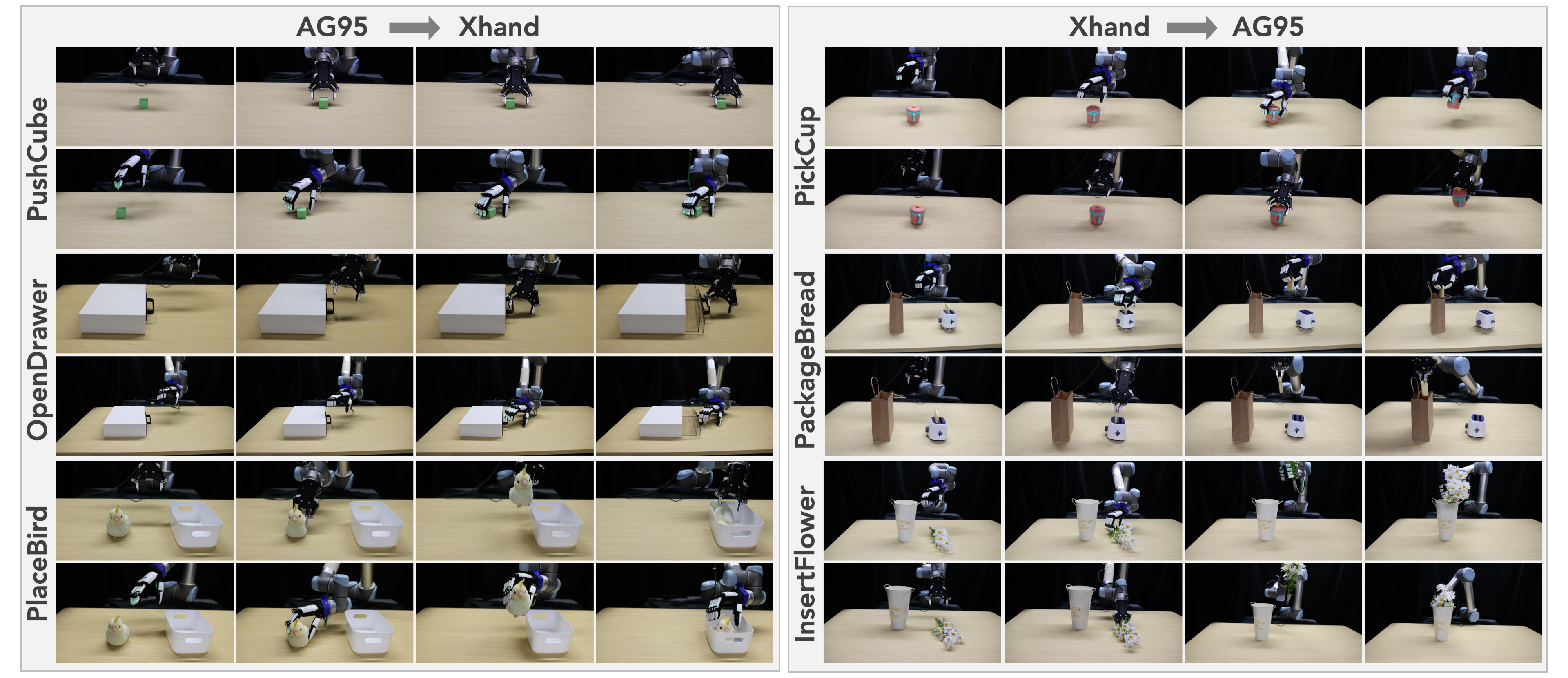}
  \caption{Qualitative evaluation. Left: Transfer from AG95 to Xhand in PushCube, OpenDrawer, and PlaceBird. Right: Transfer from Xhand to AG95 in PickCup, PackageBread, and InsertFlower. Manipulations of source policies are shown on the top rows, transferred ones on the bottom rows.}
  \label{fig:qualitative_eval}
  \vskip -0.1in
\end{figure*}

\subsubsection{Policy evaluation}

Table~\ref{tab:policy_eval} presents the policy evaluation results using synthesized cross-embodiment data. We observe that \CEI achieves an average success rate of 62\%, suggesting that the synthesized data effectively captures critical behaviors such as reaching the target position, grasping, and releasing, even without additional refinement or selection. We further compare \CEI against two baselines: (1) No Augmentation, where the policy is trained solely on observations from source demonstrations, and (2) \CEI without Inference Augmentation, where the policy uses raw observations during inference without additional augmentation. Policies trained without any augmentation fail to complete the tasks, demonstrating the necessity of targeted data augmentation for cross-embodiment generalization. Additionally, removing Inference Augmentation results in a 22\% drop in success rate. \textcolor{black}{This performance degradation arises due to the partial and noisy observations during inference.}

\begin{table}[ht]
\centering
\scriptsize
\setlength{\tabcolsep}{4pt} 

\caption{Policy Evaluation on Synthesized Data Generated by \CEI.}
\label{tab:policy_eval}
\begin{tabular}{llccccc}
\toprule
\textbf{Task} & \textbf{Method} & Emb. 1 & Emb. 2 & Emb. 3 & Emb. 4 & \textbf{Avg.} \\
\midrule

\multirow{3}{*}{\textit{OpenDrawer}}
 & No Aug.            & \(0\)   & \(0\)   & \(0\)   & \(0\)   & \(0\)   \\
 & \CEI w/o Inf. Aug. & \(\boldsymbol{100}\) & \(79\)  & \(32\)  & \(64\)  & \(69\)  \\
 & \textbf{\CEI (Ours)}   & \(\boldsymbol{100}\) & \(\boldsymbol{90}\) & \(\boldsymbol{100}\) & \(\boldsymbol{90}\) & \(\boldsymbol{95}\) \\
\midrule

\multirow{3}{*}{\textit{PickCube}}
 & No Aug.            & \(0\)   & \(0\)   & \(0\)   & \(0\)   & \(0\)   \\
 & \CEI w/o Inf. Aug. & \(42\)  & \(26\)  & \(38\)  & \(10\)  & \(29\)  \\
 & \textbf{\CEI (Ours)}   & \(\boldsymbol{79}\) & \(\boldsymbol{68}\) & \(\boldsymbol{68}\) & \(\boldsymbol{23}\) & \(\boldsymbol{60}\) \\
\midrule

\multirow{3}{*}{\textit{StackCube}}
 & No Aug.            & \(0\)   & \(0\)   & \(0\)   & \(0\)   & \(0\)   \\
 & \CEI w/o Inf. Aug. & \(36\)  & \(10\)  & \(44\)  & \(0\)   & \(23\)  \\
 & \textbf{\CEI (Ours)}   & \(\boldsymbol{55}\) & \(\boldsymbol{16}\) & \(\boldsymbol{52}\) & \(0\)   & \(\boldsymbol{31}\) \\
\midrule

\multicolumn{2}{l}{\textcolor{gray}{Source (ref.)}} 
 & \multicolumn{4}{c}{\textcolor{gray}{100 / 100 / 97}} 
 & \textcolor{gray}{99} \\

\bottomrule
\end{tabular}
\end{table}

\subsection{Real-world Results}

\subsubsection{Bidirectional transfer}

Table~\ref{tab:realworld} demonstrates the bidirectional transfer capabilities of \CEI on real-world tasks. We compare policies trained on synthesized data and deployed on the target embodiment against those trained on source data and deployed on the source embodiment. Results show that for simple tasks such as PushCube, \CEI enables transfer from a parallel-jaw gripper to a dexterous hand without performance loss. Failures in tasks like OpenDrawer and PlaceBird are primarily due to challenges in dexterous contacts, such as fingers slipping off drawer handles. In transfers from a dexterous hand to a parallel gripper, \CEI achieves similar performance. \textcolor{black}{However, InsertFlower remains exceptionally challenging since the thin geometry of the flower stem makes it prone to slippage, leading to low success rates for both source and target embodiments.} Overall, \CEI achieves an average success rate of 70\% across six tasks, with a transfer ratio (success rate of \CEI divided by that of the source embodiment) of 82.4\%. Qualitative evaluation across 6 tasks is shown in Fig.~\ref{fig:qualitative_eval}.

\begin{table}[ht]
\centering
\scriptsize
\caption{Real-world Evaluation.}
\label{tab:realworld}
\begin{tabular}{lccc|c}
\toprule
\multicolumn{5}{c}{\textbf{AG95 $\rightarrow$ Xhand}} \\
\midrule
\textbf{Method} & PushCube & OpenDrawer & PlaceBird & Average \\
\midrule
\textbf{\CEI (Ours)} & $10/10$ & $8/10$ & $7/10$ & $8.3/10$ \\
Source & $10/10$ & $10/10$ & $10/10$ & $10/10$ \\
\midrule
\multicolumn{5}{c}{\textbf{Xhand $\rightarrow$ AG95}} \\
\midrule
\textbf{Method} & PickCup & PackageBread & InsertFlower & Average \\
\midrule
\textbf{\CEI (Ours)} & $6/10$ & $9/10$ & $2/10$ & $5.7/10$ \\
Source & $8/10$ & $9/10$ & $4/10$ & $7/10$ \\
\bottomrule
\end{tabular}
\end{table}

\begin{figure*}[t]
  \centering
  \vspace{2pt}
  \includegraphics[width=.95\linewidth]{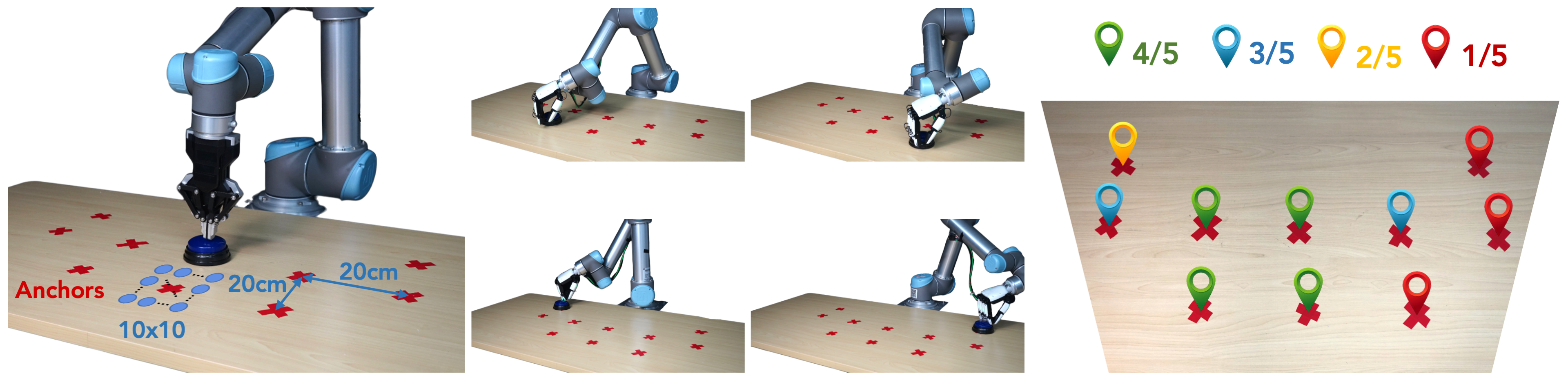}
  \caption{Spatial generalization. Left: Experimental setup and task configuration, where each anchor is spaced 20\,cm apart. Around each anchor, we sample a $10 \times 10$ grid within the range $[-8\,\text{cm}, 8\,\text{cm}]$ for data generation. Middle: Our approach enables the Xhand to press the button across most of the table surface. Right: Evaluation results on the 10 anchors. Each anchor is evaluated over 5 trials, with the results distinguished by different colors.}
  \label{fig:spatial generalization}
\vskip -0.2in
\end{figure*}

\subsubsection{Time cost of transfer}

To study how efficient the data generation process is, we evaluate the efficiency of \CEI’s data generation by measuring the time required to synthesize 100, 200, 300 and 400 demonstrations. Given the substantial cost of collecting large-scale real-world demonstrations, we construct larger datasets by replicating a set of 25 collected trajectories through duplication at ratios of 4$\times$, 8$\times$, 12$\times$, and 16$\times$, respectively. We compare \CEI with MimicGen~\cite{pmlr-v229-mandlekar23a} and DemoGen~\cite{xue2025demogen} on datasets with an average episode length of 105 steps. For MimicGen, we estimate time by multiplying the replay duration of each source trajectory by the number of generated demonstrations, and adding 20 seconds per trajectory for manual object resets, following~\cite{xue2025demogen}. Table~\ref{tab:time_cost} presents the time required to generate those numbers of demonstrations. Results show that \CEI requires significantly less time than MimicGen, which highly depends on online execution. DemoGen generates hundreds of demonstrations in one second, while \CEI requires several minutes since it utilizes gradient-based optimization. 

\begin{table}[ht]
\centering
\caption{Time Cost for Generating Real-world Demonstrations.}
\scriptsize
\label{tab:time_cost}
\begin{tabular}{c|cccc}
\toprule
& 100 Demos & 200 Demos & 300 Demos & 400 Demos\\
\midrule
MimicGen & $6.4$ h & $12.8$ h & $19.2$ h & $25.6$ h\\
DemoGen & $0.29$ s & $0.54$ s & $0.84$ s & $1.14$ s\\
\textbf{\CEI (Ours)} & $2.5$ min & $2.9$ min & $3.3$ min & $3.6$ min\\
\bottomrule
\end{tabular}
\end{table}

\vspace{-0.5cm}

\section{Broader Applications}

\subsection{Spatial Generalization}

\textcolor{black}{
\subsubsection{Spatial augmentation} 
\CEI inherently facilitates spatial generalization, enabling the generation of diverse demonstrations across the entire workspace from only a single collected trajectory. Given a spatial transform $\mathcal{T}_i$ which encapsulates the possible translational or rotational offset of the objects, we first apply it to the functional representation trajectory:
\begin{equation}
    \tilde{X}_t = X_t + \mathcal{L}(t)(\mathcal{T}_i(X_t) - X_t), \quad t = 0,1,\ldots,L-1,
\end{equation}
where \( \mathcal{L}(t) = \min(\frac{t}{0.8L}, 1)\) is the clipped linear growth, ensuring that the generated demonstrations share an initial state while diverging to different terminal states by interpolation. The target embodiment is subsequently aligned to the augmented trajectory 
$\{\tilde{X}_t\}_{t=0}^{L-1}$ 
through the standard \CEI optimization procedure. 
The augmented point cloud is then obtained by applying $\mathcal{T}_i$ to the object point cloud and synthesizing the robot point cloud according to the augmented trajectory.
}
\subsubsection{Press the button anywhere on the table}
We assess the spatial generalization of \CEI in the PressButton task. Starting from a single demonstration collected with AG95, we generate 1{,}000 demonstrations across 10 anchors, where each anchor samples a $10 \times 10$ grid within the range $[-8\,\text{cm}, 8\,\text{cm}]$ (Fig.~\ref{fig:spatial generalization}, left). We then train DP3 and evaluate it on each anchor position for 5 trials. As illustrated in Fig.~\ref{fig:spatial generalization} (middle and right), our approach extends the policy to press the button over a wide area of the table, rather than being limited to the original position. Moreover, we observe that performance on the left side is better than on the right, likely because the camera is positioned to the right of center and oriented toward the left.

\subsection{Multimodal Motion Generation}

\begin{figure}[t]
  \centering
  \includegraphics[width=.9\linewidth]{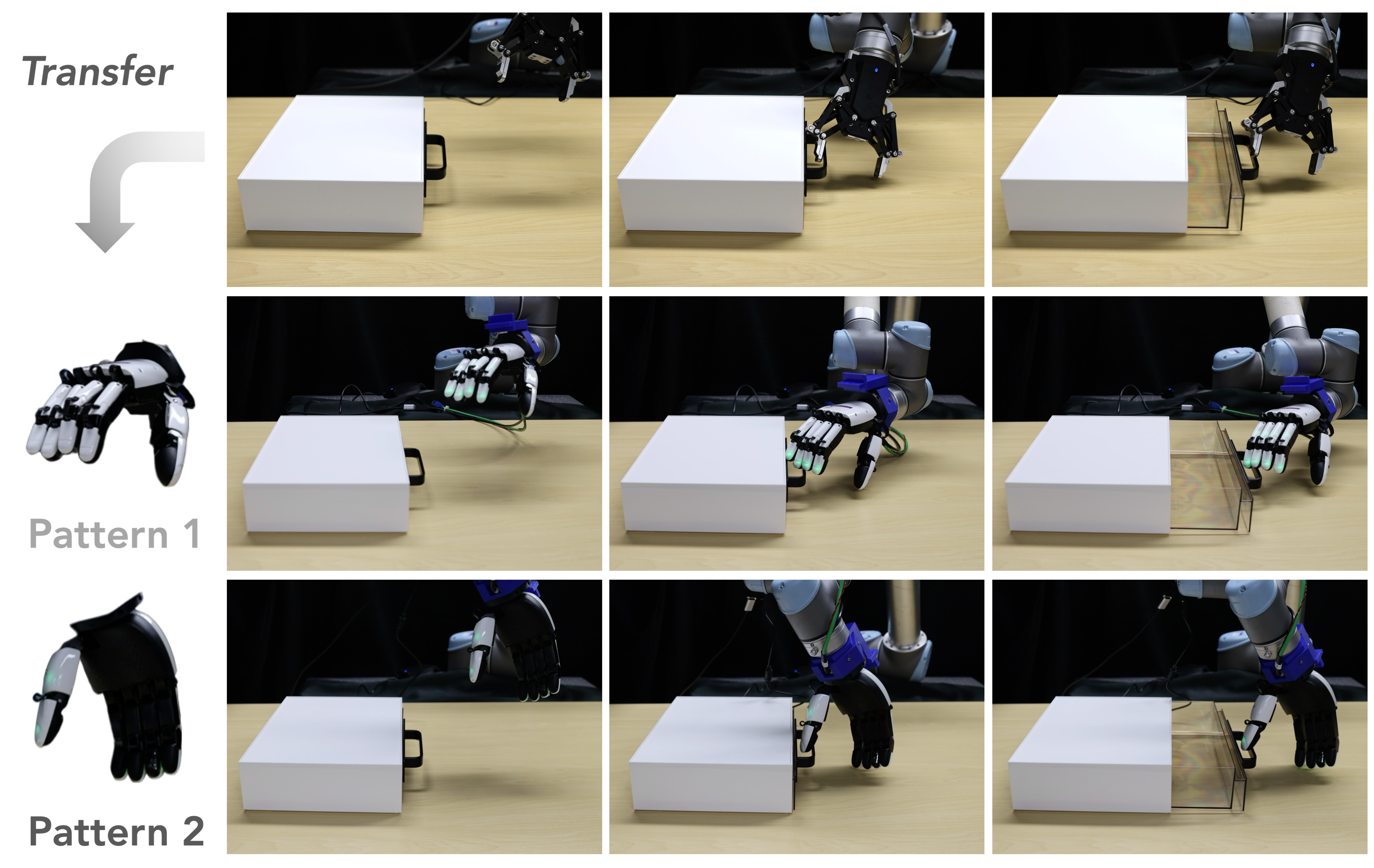}
  \caption{Multimodality of \CEI. \CEI generates two different manipulation motions in OpenDrawer task from the same demonstration.}
  \label{fig:multimodality}
\vskip -0.1in
\end{figure}

In previous experiments, we observed that varying the initial joint configuration produces different aligned trajectories. For example, when transferring a grasp motion from a gripper to a dexterous hand, the resulting pose of the dexterous hand remains valid even if rotated by $180^{\circ}$ around the heading direction. To exploit such multimodality, we introduce an initialization strategy designed for enhancing data diversity.

\subsubsection{Elite-based initialization strategy (EIS)} 
 Instead of relying on random joint configurations, we uniformly sample candidate configurations from the joint space and rank them according to their \textit{functional similarity} with the functional representation of the source embodiment’s initial configuration. The top 10\% of these candidates are then selected, and their mean configuration is used as the initialization. Subsequently, we proceed with the original process of \CEI to obtain the aligned trajectory and synthesize corresponding demonstrations.

\subsubsection{Bimodal motion in OpenDrawer}
We evaluate the extension in OpenDrawer task. As shown in Fig.~\ref{fig:multimodality}, \CEI generates 2 different patterns that successfully open the drawer. Since we only manipulate the initialization and \CEI proceeds the trajectory sequentially, the aligned trajectories remain temporally consistent while allowing for diverse motion patterns. We further train DP3 on 25 demonstrations with either an arbitrary motion pattern or a 1:1 mixture of patterns. As reported in Table~\ref{tab:open_drawer}, the multimodality has no adverse impact on task success.

\begin{table}[ht]
\centering
\caption{Success Rates of Different Training Recipes.}
\label{tab:open_drawer}
\begin{tabular}{cccc}
\toprule
   & Pattern 1 & Pattern 2 & Mixture of Pattern 1\&2 \\
\midrule
Success Rate & \(8/10\) & \(7/10\) & \(7/10\)\\
\bottomrule
\end{tabular}
\end{table}

\vspace{-0.3cm}
\section{Conclusion and Discussion}
\label{sec:conclusion}

In this letter, we introduce \CEI, a cross-embodiment framework that leverages \textit{functional similarity} and automated data synthesis to transfer policies across diverse robots in both simulation and the real world. Beyond standard manipulation, \CEI supports spatial generalization and multimodal motion generation, providing a versatile foundation for scalable robot learning. While our current scope is point cloud-based cross-embodiment learning, preliminary results suggest our work also holds potential for RGB-based observations, as shown on our project website \url{https://cross-embodiment-interface.github.io/}. Although \CEI can synthesize hundreds of demonstrations in parallel within minutes, applying it to large-scale datasets remains an open direction that could further advance generalist policy learning. In addition, the current reliance on visual-kinematic inputs limits the detection of unstable contacts, such as the slippage seen in the InsertFlower task. Integrating tactile sensing would allow the policy to adapt to these physical disturbances in real-time, representing a key area for future improvement.

\bibliographystyle{ieeetr}
\bibliography{main}
\end{document}